\documentclass[journal,transmag]{IEEEtran}
\ifCLASSINFOpdf
    \usepackage[utf8]{inputenc}
    \usepackage{graphicx}
    \usepackage{amssymb}
    \usepackage{algorithm}
    \usepackage{listings}
    \usepackage{etoolbox}
    \usepackage{algorithmic}
    \usepackage{xcolor}
    \usepackage{bm}
    \usepackage{url}
    \usepackage{quoting, lipsum}
    \usepackage{float}
    \usepackage{hyperref}
    \usepackage{cleveref}
    \crefname{section}{§}{§§}
    \Crefname{section}{§}{§§}
\else
\fi

\begin{document}

\title{Self-supervised Temporal Discriminative Learning for Video Representation Learning}




\author{\IEEEauthorblockN{Jinpeng Wang\IEEEauthorrefmark{1}\textsuperscript{\textsection},
~\IEEEmembership{Student Member,~IEEE},
Yiqi Lin \IEEEauthorrefmark{2}\textsuperscript{\textsection},
Andy J. Ma\IEEEauthorrefmark{2}, 
and
Pong C. Yuen \IEEEauthorrefmark{3},~\IEEEmembership{Senior Member,~IEEE}}

\IEEEauthorblockA{\IEEEauthorrefmark{1}School of Electronics and Information Technology, Sun Yat-sen University, Guangzhou, 510006, China}
\IEEEauthorblockA{\IEEEauthorrefmark{2}School of Data and Computer Science, Sun Yat-sen University, Guangzhou, 510006, China}
\IEEEauthorblockA{\IEEEauthorrefmark{3}
Department of Computer Science, Hong Kong Baptist University, Hong Kong}

\thanks{
Corresponding author: Andy~J. Ma (email: majh8@mail.sysu.edu.cn).}}

\markboth{Self-Supervised Learning Using Consistency Regularization of Spatio-Temporal Data Augmentation for Action Recognition}%
{Shell \MakeLowercase{\textit{et al.}}: Bare Demo of IEEEtran.cls for IEEE Transactions on Magnetics Journals}
%

\IEEEtitleabstractindextext{%
\begin{abstract}
Temporal cues in videos provide important information for recognizing actions accurately. 
However, temporal-discriminative features can hardly be extracted without using an annotated large-scale video action dataset for training.
This paper proposes a novel Video-based Temporal-Discriminative Learning (VTDL) framework in self-supervised manner.
Without labelled data for network pretraining, temporal triplet is generated for each anchor video by using segment of the same or different time interval so as to enhance the capacity for temporal feature representation.
Measuring temporal information by time derivative, Temporal Consistent Augmentation (TCA) is designed to ensure that the time derivative (in any order) of the augmented positive is invariant except for a scaling constant.
Finally, temporal-discriminative features are learnt by minimizing the distance between each anchor and its augmented positive, while the distance between each anchor and its augmented negative as well as other videos saved in the memory bank is maximized to enrich the representation diversity.
In the downstream action recognition task, the proposed method significantly outperforms existing related works.
Surprisingly, the proposed self-supervised approach is better than fully-supervised methods on UCF101 and HMDB51 when a small-scale video dataset (with only thousands of videos) is used for pre-training.
The code has been made publicly available on \href{https://github.com/FingerRec/Self-Supervised-Temporal-Discriminative-Representation-Learning-for-Video-Action-Recognition}{\color{blue}{https://github.com/FingerRec/VTDL}}.
\end{abstract}


\begin{IEEEkeywords}
self-supervised learning, temporal triplet, temporal consistent augmentation,  temporal-discriminative representation learning.
\end{IEEEkeywords}}

\maketitle

\IEEEdisplaynontitleabstractindextext

\IEEEpeerreviewmaketitle

\begin{figure*} 
  \centering
  \includegraphics[width=.8\linewidth]{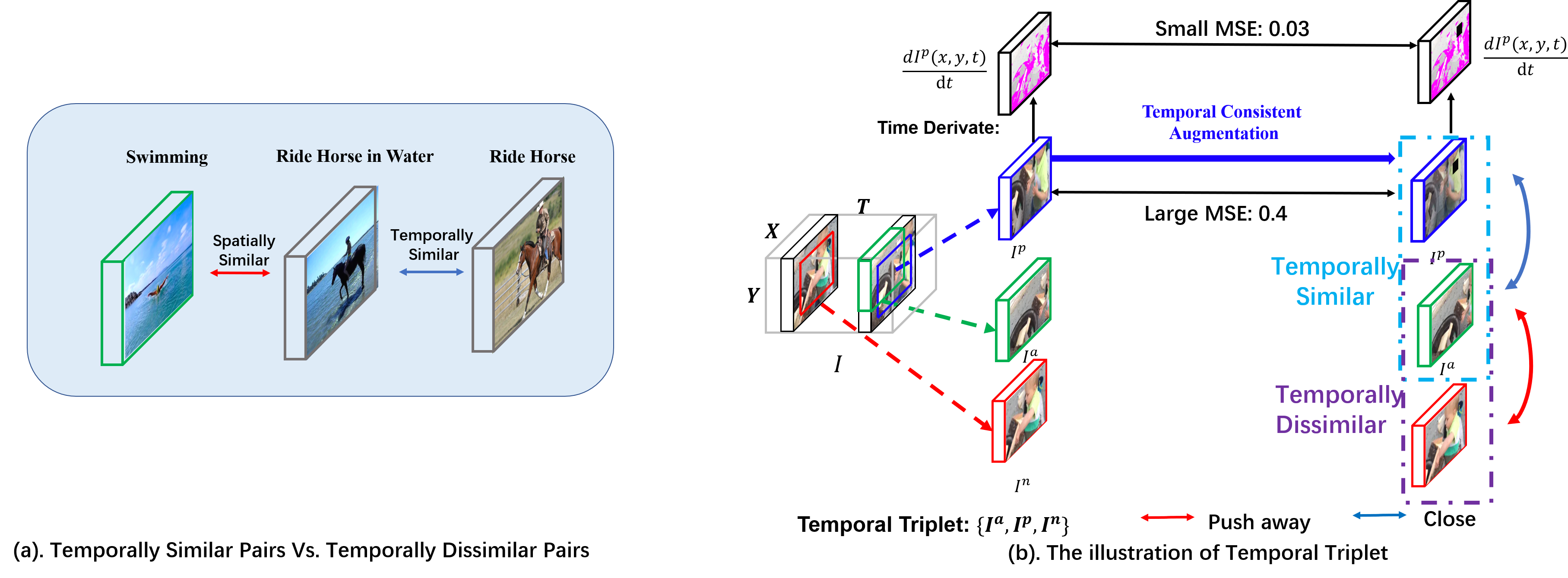}
  \caption{
  (a) Illustration of temporally similar and dissimilar pairs.
  (b) Illustration of the data augmentation idea in the proposed temporal-discriminative representation learning method. 
  In our work, temporal triplet is generated based on the observation that different time intervals of a video clip can be considered as distinct instances with different feature representations. 
  Measuring the temporal similarity by any-order time derivative, the proposed Temporal Consistent Augmentation (TCA) reduces the spatial similarity of the positive pair, but maintains the temporal invariance except for a scaling constant.
  }
  \label{fig:teaser}
\end{figure*}
\maketitle

\begingroup\renewcommand\thefootnote{\textsection}
\footnotetext{Equal contribution}

\section{Introduction}
As the usage of sensor-rich mobile devices increases rapidly, video plays an important role as one of the communication carriers.
To develop advanced video-understanding techniques, action recognition become a fundamental issue.
As an information-intensive multimedia, the additional time dimension in videos make the task of action recognition more challenging compared with image classification.
With the help of 3D convolutional neural networks (CNNs) \cite{carreira2017quo,wang2018non,feichtenhofer2019slowfast} and large-scale manually annotated video datasets, it has achieved remarkable progress to learn discriminative spatio-temporal representations from labeled videos for action recognition in recent years.
However, as video annotation requires additional temporal segmentation, it is much more expensive and time-consuming to annotate a large number of videos manually than image. 

Without a large-scale annotated dataset for training, self-super-vised learning has been proposed recently to provide human-labor-free supervision signal by defining a pretext task from a set of unlabelled data for pretraining .
In image applications, pretext tasks in existing methods can be defined by predicting the transformations \cite{gidaris2018unsupervised}, the patch orders \cite{DBLP:conf/iccv/DoerschGE15,DBLP:conf/eccv/NorooziF16}, the missing information \cite{DBLP:conf/cvpr/PathakKDDE16} and so on.
Instead of random initialization, by optimizing for these pretext tasks, useful semantic features are extracted for down-stream vision tasks.
Though it has achieved great success in existing self-supervised image representation learning methods \cite{wu2018unsupervised,ye2019unsupervised,ZhuangZY19,he2019momentum, chen2020simple}, directly employing these methods for videos may not make good use of the temporal information which is very important in video applications. 


With the extra time dimension in videos, several related pretext tasks are proposed by using the temporal charerterastics, such as the frame order verification \cite{misra2016shuffle}, frame order prediction \cite{xu2019self}, spatio-temporal cubic puzzles \cite{kim2019self}, etc.
While most of these existing methods are developed by predicting or reconstructing the temporal or spatio-temporal order of frames or spatio-temporal cubics, temporal similarity as illustrated in Figure \ref{fig:teaser}(a) has hardly been considered in their representation learning models.
Without measuring temporal similarity, it may be impossible to know whether the learnt features are temporally discriminative or not.

To overcome the problems in existing methods, we propose a novel temporal-discriminative representation learning method for self-supervised video action recognition.
Without labelled data for network pre-training, the proposed method first generates temporal  similar and dissimilar data to learning discriminative features in time dimension. 
The idea of the data augmentation step is shown in Figure \ref{fig:teaser}(b).
Since video segments in different time intervals of a video clip can be considered as distinct instances, temporal triplet is generated by sampling temporally similar and dissimilar pairs as positive and negative, respectively, with respect to an anchor video segment.
For the positive of an anchor, both the temporal and spatial similarities between the positive and the anchor are high as then are in the same time interval of the same video clip.
In this case, by extracting only the spatially discriminative features (without temporally discriminative features), the positive could already be close to the anchor.
To ensure that temporally discriminative features can be learnt, a further data augmentation method namely Temporal Consistent Augmentation (TCA) is proposed by maintaining the time derivative in any order invariant except for a scaling constant. 
As a result, the positive in a temporal triplet after TCA is temporally similar but spatially dissimilar to the anchor.


With the generated data, the proposed Video-based Temporal-Discriminative Learning (VTDL) maximizes the distance of the negative pair and minimizes the distance of the positive pair after TCA.
To enrich the diversity for feature representation, different samples (inter-video) saved in the memory bank are also treated as negatives temporally dissimilar to the anchor in the embedding space.
In summary, the contributions of this work are  as follows:
\begin{itemize}
    \item We propose a novel self-supervised learning framework namely Video-based Temporal-Discriminative Learning (VT-DL) for action recognition. 
    \item To learn temporal-discriminative representation, a temporal triplet generation method is proposed based on the observation that different time intervals of a video clip can be considered as distinct instances with different feature representations.
    \item To ensure temporal discriminative features (instead of spatially discriminative ones) can be extracted, a Temporal Consistent Augmentation (TCA) method is proposed to reduce the spatial similarity of the positive pair. 
    Measuring the temporal similarity by any-order time derivative, the proposed TCA maintains temporal invariance except for a scaling constant.
    \item Extensive experimental results show that the proposed VTDL significantly outperforms the state-of-the-arts on publicly available benchmarks.
\end{itemize}

\section{Related Work}
\begin{figure*}
  \includegraphics[width=1\linewidth]{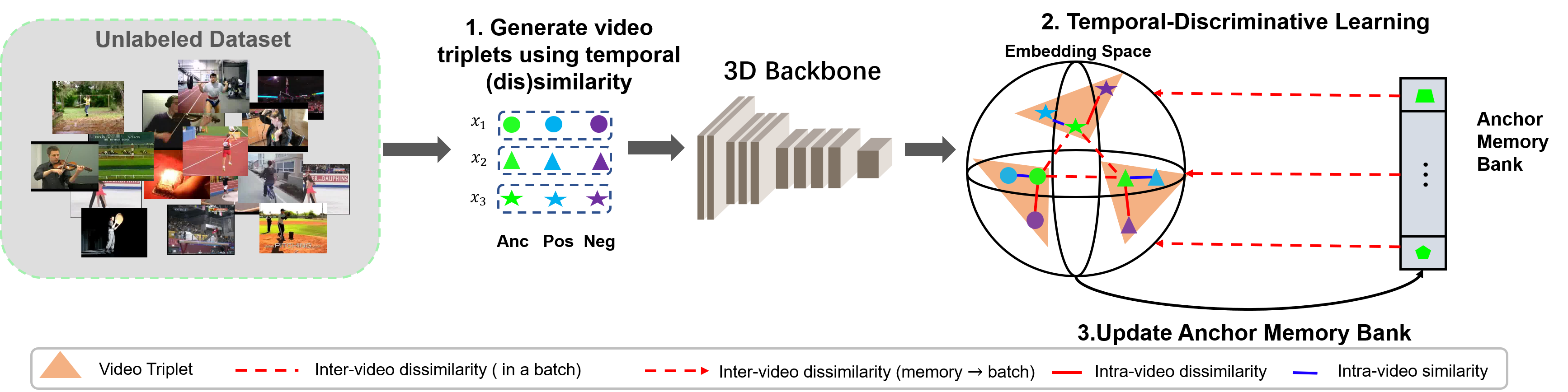}
  \caption{Framework of the proposed \emph{Video-based Temporal-Discriminative Learning}, which includes three steps in one iteration: 
  Step 1) Generating temporal triplets for each video in a training batch;
  Step 2) Self-supervised learning with Temporal-Discriminative Loss for temporal-discriminative feature extraction with a 3D Backbone Network;
  Step 3) Updating Anchor Memory Bank with anchor features in each training batch. (Lines in red denote dissimilarity while lines in blue denote similarity. This figure is better viewed in color.)
  }
  \label{fig:freamwork}
\end{figure*}
In this section, we first introduce self-supervised representation learning in image and video domain in \cref{sec:ssl}, then we present a brief review of video action recognition in \cref{sec:action_recognition}.

\subsection{Self-supervised Representation Learning}
\label{sec:ssl}
Obtaining labeled data is very expensive while plentiful unlabeled data are readily available. 
As a prominent unsupervised representation learning paradigm, self-supervised learning have drawn a lot of attention in these years.
In image domain, a lot of works try to learn semantic information with predefined surrogate signal.
For example, \cite{gidaris2018unsupervised} utilizing the rotation degree of image as surrogate and \cite{doersch2015unsupervised} reasoning the relative positions between two image patches. 
Another series of work \cite{wu2018unsupervised,ye2019unsupervised,ZhuangZY19,he2019momentum, chen2020simple} based on contrastive learning and powerful data augmentation achieve remarkable success on unsupervised representation learning.
Contrastive learning \cite{hadsell2006dimensionality} aims at learning an instance embedding space by measuring the similarities of sample pairs. 

Recently, the research interest has been extended to video domain.
A majority of these works take advantage of the temporal structure in videos. 
Early works usually learn temporal information based on 2D CNN and focus on frame-by-frame relation, such as verification of order \cite{misra2016shuffle}, predicting order \cite{fernando2017self}.
Recently, with the development of 3D CNNs, the pretext tasks focus more on spatio-temporal modeling, such as keeping color-consistency \cite{wang2019self} across frames or predicting the clip order \cite{xu2019self} and solving space-time cubic puzzles \cite{kim2019self}.
Another recently introduced novel objective function is feature prediction, such as dense predictive coding \cite{han2019video} and video cloze procedure \cite{luo2020video}. 
Our work is closely related to TCN \cite{sermanet2018time}, which first introduce contrastive learning in robotic dataset. 
However, effective multi-view TCN generate positive pairs from multiple viewpoints captured by robotic, which is not suitable in video dataset. 
Although single-view TCN can be extend to conventional video by generating positive pairs in a small temporal gap around anchor, single-view TCN has very weak performance even behind Shuffle \& Learn \cite{misra2016shuffle}.

Although these methods can handle temporal information by solving the pretext tasks, the learned representation may lack of temporal-discriminative ability.
We address this problems by introducing Data Augmentation, an important but undervalued topic in video self-supervised learning.
Especially, Our proposed Temporal Consistent Augmentation also can be a practical solution for learning temporal-discriminative video representation in different supervision paradigm.

\subsection{Video Action Recognition}
\label{sec:action_recognition}
Video Action Recognition is a active area of vision research and a variety of 2D and 3D methods have been proposed \cite{wang2018temporal}. 
Observing that 2D CNNs cannot directly exploit the temporal patterns of actions, the 3D CNNs become more popular recently \cite{carreira2017quo,feichtenhofer2019slowfast, DBLP:conf/mm/RuizPBM17,DBLP:journals/tmm/HouWSJ18,DBLP:conf/mm/JiX00SH19,DBLP:conf/mm/LiuGQWL19,DBLP:conf/mm/0003ZWCLC19}.
These works stack 3D convolutions to jointly model temporal and spatial semantics.
With the introduction of large action recognition datasets (e.g., Kinetics), higher accurate and deeper 3D video CNNs are possible to reach \cite{wang2018non,hara2018can,tran2018closer}.
However, the massive parameters of 3D CNNs make the initialization become a serious problem.
The common practice is either random initialize \cite{hara2018can} or inflated from 2D ImageNet pretrained weight \cite{carreira2017quo}, which show very modest performance.
Another option is fully-supervised learning with large video datasets, but the annotation and pretraining are very time-consuming and expensive.
In this work, we evaluate our proposed method on four widely used 3D CNNs and show that they can be further improved by unsupervised pretrain.
As a result, our method is an alternative option for video CNN pretrain and more in line with practical scenarios.

\section{Video-base Temporal-Discriminative Learning}
In this section, we first present framework overview.
Then we describe generating strategy and data augmentation used in temporal triplet. Finally, we introduce the temporal-discriminative learning.

\subsection{Framework Overview}
\label{sec:framework}
The overall architecture of our method is showed in Figure \ref{fig:freamwork}.
First, for each video $x_i$ in the unlabeled dataset $D$ we generate temporal triplet with novel augmentation by the generator $g$.
Next, we use the penultimate later of network $f$ to extract features from each generated temporal triplet and then project the high-level feature maps into fix-length vector.
Based on these vectors, we use temporal-discriminative loss to optimize $f$ and save anchor feature of temporal triplet into the memory bank.
All these steps are iteratively performed until convergence in completely unsupervised manner.

\begin{figure*} 
  \centering
  \includegraphics[width=.9\linewidth]{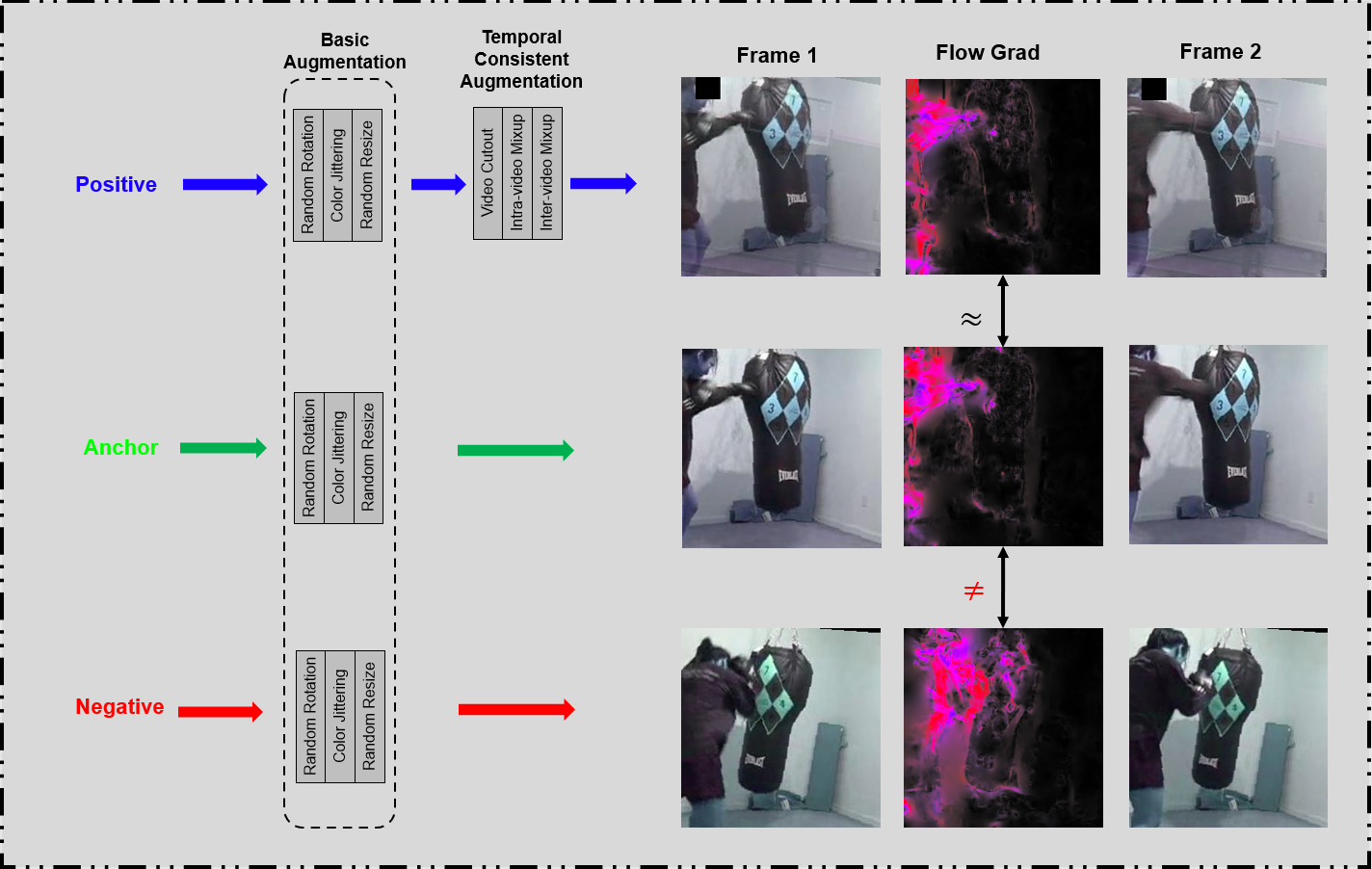}
  \caption{The pipeline of data augmentation for Temporal Triplet generator.
  The flow grad of visualized samples shows that the positive pairs share similar temporal cue while a big difference on pixel-level.}
  \label{fig:triplet}
\end{figure*}

\subsection{Generating Temporal Triplet.}
To generate temporal similar/dissimilar pairs, we introduce the concept of triplet \cite{schroff2015facenet} into our framework and propose video-based Temporal Triplet.
Each Temporal Triplet consist of anchor, positive, and negative and all of them are 3D spatio-temporal pieces cropped from a video.
Specifically, for each video $x_i$ in $D$, we first randomly crop one 3D spatio-temporal piece, named as Anchor. 

\textbf{Positive:}
One possible technique to sample temporal similar pair is treating another spatio-temporal piece temporally nearby anchor as positive like TCN \cite{sermanet2018time}.
However, the temporal information may not stay similar enough between  anchor and positive even with pretty close timestamps, when intense changes take place within a short time, for example, archery.
To avoid such ambiguity, the positive is cropped from the same temporal but different spatial location as anchor, which make sure that positive pairs can stay temporally similar but be different in pixel-level.

\textbf{Negative:}
In a video, negative are sampled from different temporal interval with anchor but no requirement for spatial location.
The anchor and negative at least vary constant $\tau$ timesteps.
Formally,
\begin{equation}
\label{eq:2}
    |t_a - t_n| > \tau
\end{equation}
where $t_a$ and $t_n$ is the begin timestamp of anchor and negative.
Intuitively, when the timestamp between anchor and negative is close, the base model is encouraged to focus more on fine-grained temporal information, since they have a high chance to share similar even same spatial context.
While with large timestamp gap, the network may easily separate them apart by different spatial object or context, which will damage the temporal-discriminative representation learning.
Therefore, the choice of $\tau$ is quite important and the detail experiment is given in \cref{sec:ablation_study}.


\textbf{Basic Augmentation:}
Unfortunately, only generating temporal similar/dissimilar pairs via above sampling strategy may result in the base model easily distinguish temporal triple by low-level trivial solution (such as Color, Edges, Objects).
In order to overcome this obstacle, Basic Augmentation (BA) is first adopted to bring large visual variation in pixel-level for each temporal triplet.
Basic augmentation is type of transformation such as scale, occlusion, lighting, rotation or any combination of them. 
In this paper, the basic augmentation setting is: A $224 \times 224$ pixel cropping is taken in a randomly resized video, and then undergoes color jittering, random rotation. 
As the flip or rotation with huge degree may confuse the visual temporal cue (e.g., sit and stand up), we rotate the overall video with small degree.
In practice, the rotation not surpass 10 degree is recommended.





\subsection{Temporal Consistent Augmentation}

To diversify the positives, the temporal information between origin and augmented should keep similar under same measurement, time derivative.
Inspired by the time derivative invariant, we further proposed Temporal Consistent Augmentation.

\subsubsection{Time derivative invariant}
The temporal information is ambiguous and hard to define.
In early hand-craft video features \cite{1544882,5196739,910878}, the inter-frame differences are used to provide useful indication of motion.
Along this line, time derivative can be used to measure the variation in temporal information.
In particular, video can be considered as a space-time function $I(x,y,t)$,the $x$ and $y$ are the pixel index of each frame, $t$ is the time dimension.
It is easily to prove that the time derivative stay consistent in any order when addition or multiplication operations with constant are applied into space-time function.
Formally,
\begin{equation} 
     \frac{ \mathrm{d^{k}}(\alpha I(x,y,t)+(1-\alpha)\delta(x, y))}{\mathrm{d}t^{k}} = \alpha \frac{\mathrm{d^{k}}I(x,y,t)}{\mathrm{d}t^{k}}
\end{equation}
where $\alpha$ and $\delta$ denote a constant scaling factor of multiplication operation and a function only depends on $x$ and $y$ for  addition operation.
Therefore, the extra spatial context (image) can be introduced into the space-time function (video) while time derivative stay consistent in any order with a scaling factor.
In other words, the $\delta$ can be a static image and the whole operation can be viewed as mixing up same image into every frame of video. 
When an appropriate scaling factor is chosen, the temporal cues can stay similar under dissimilar spatial context.
Furthermore, the optical flow measurements of temporal triplet in Figure \ref{fig:triplet} can validate our assumption.

\subsubsection{Temporal Consistent Augmentation}
Based on above analysis, Temporal Consistent Augmentation introduce video cutout regularization and mixing up image into every frame of positive to vary spatial content in pixels while keep temporal pattern similar.
Considering an video $x_i$ in $D$ with length $L$, we define 0-1 mask $\mathcal{M}$ and other image frame $\mathcal{N}$.
The Temporal Consistent Augmentation can be formalized as:
\begin{equation}
\label{eq:3}
    \hat{x}_i^j = (\alpha x_i^j  + (1 - \alpha) \mathcal{N})\odot \mathcal{M}, j \in [1, L]
\end{equation}
where $\alpha$ is uniform sampled from $[0.5,1]$ and $x_i^j$ is the $j$th frame of $x_i$.
Both $\mathcal{M}$ and $\mathcal{N}$ have same spatial size as $x_i^0$.

Based on this formulation, we propose three instantiations for Temporal Consistent Augmentation:
\emph{i.} Video Cutout: taking inspiration from \cite{DBLP:journals/corr/cutout}, we randomly cutout one region by zeroing the $\mathcal{M}$ at the selected region and set $\alpha$ to 1.
\emph{ii.} Internal Mix: we set $\mathcal{M}$ to 1 for all elements and select one random frame from $x_i$ as $\mathcal{N}$. 
\emph{iii.} External Mix: we set $\mathcal{M}$ to 1 for all elements and select $\mathcal{N}$ from another video $x_j, j \neq i$.

Selecting mixed frame from inter- and intra- sample aims at enriching the diversity of spatial context.
In this paper, our Temporal Consistent Augmentation is in a cascade of these three data augmentations. 
The pipeline of data augmentation used in temporal triplet and visualized samples are illustrated in Figure \ref{fig:triplet}.

\subsection{Temporal-Discriminative Learning}
\subsubsection{Temporal-Discriminative Loss.}
Inspired by the contrastive Loss \cite{hadsell2006dimensionality} and triplet loss \cite{schroff2015facenet,weinberger2006distance} from metric learning, we propose temporal-discriminative loss to better leverage the online generated temporal triplet.
It is based on a naive idea that the feature of temporally similar samples should keep closer to each other.
Suppose we have an unlabeled dataset $D$ with $N$ samples $\{x_1,...,x_N\}$, their high-level representation $\{v_1,...,v_N\}$ are generated by the embedding function $v_i=f_\theta(x_i)$, where
$f_{\theta}$ is a deep network with parameter $\theta$, mapping video $x$ to feature vector $v$.
The temporal-discriminative loss can be written as:
\begin{equation}
    \label{eq:4}
    L = \sum_{i=1}^{N} - log \frac{d(v_i^a, v_i^p) }{d(v_i^a, v_i^p) + d(v_i^a, v_i^n) + \sum_{j \neq i}^{} d(v_i^a, v_j^a)}
\end{equation}
where $d$ is a similarity measure function, which can be implemented as euclidean distance or inner product, and $\{v_i^a, v_i^p, v_i^n\}$ denote the feature vectors of temporal triplet sampled from video $x_i$.
This loss makes sure that, given a temporal triplet $\{x_i^a, x_i^p, x_i^n\}$ generated from same video sample $x_i$, the anchor’s projection $v_i^a$ is closer to the positive point’s projection $v_i^p$ than the negative one $v_i^n$ (intra-video) or $v_j^a$ (inter-video).
Note that the anchor points from different sample are also regarded as  negative pair, which enrich the diversity of temporally dissimilar pairs.
Only using the anchors of distinct sample to construct inter-video contrast is a trade-off between computational overhead and optimization.
In this way, the intra-video and inter-video negative pairs encourage the network to pay more attention to temporal content.

\subsubsection{Memory Bank.}
From equation \ref{eq:4}, all anchor features should be available at any time to generate negative pairs for temporal discriminative learning. 
However, as $N$ become larger, saving all anchor features into memory is expensive and not practical.
One possible alternative is computing temporal-discriminative loss in each mini-batch as in \cite{ye2019unsupervised}, but this technique highly relies on graphics memory, which is not feasible for video. 
Another option is using memory bank with fixed size to save features \cite{he2019momentum}, which is considered in this paper.

For each iteration, we generate triplets for each mini-batch and treat all these features saved in the memory bank as negative. 
And then we compute temporal-discriminative loss as Eq. \ref{eq:4} and update the backbone network by back-propagation.
After that, we replace the earliest feature in memory bank with anchor features in current mini-batch.
Unfortunately, in this way the anchor features saved in memory are generate via inconsistent parameter, which is instability and may bring noise during training. 
One reasonable technique to ease this problem is weight-averaged updating the network \cite{tarvainen2017mean, he2019momentum}.
Inspired by these methods, we compute negative and positive with newest network but compute anchor with moment averaged network.
The main algorithm is illustrated at Algorithm \ref{alg:1}.

\begin{algorithm}[!t]
\caption{\label{alg:1} Video-based Temporal-Discriminative Learning.}
\begin{algorithmic}
    \STATE \textbf{input:} unlabeled dataset with size $N$, network parameter $f_{\theta_t}$, history network parameter $f_{\theta_{t'}}$, smoothing coefficient hyperparameter $m$, temporal triplet generator $g$, distance measure function $d$, anchor memory bank $B$ with size $K$.
    \STATE \textbf{Initialize:} {Random Initialization $B$}
    \FOR{each sample $\{\bm x_k\}_{k=1}^N$}
        \STATE $~~~~$\textcolor{gray}{\# generate triplet} 
        \STATE $~~~~$$x_k^a, x_k^p, x_k^n = g(x_k)$ 
        \STATE $~~~~$\textcolor{gray}{\# compute anchor feature with history network}
        \STATE $~~~~$$v_k^a = f_{\theta_{t'}}(x_k^a)$ 
        \STATE $~~~~$$v_k^p = f_{\theta_t}(x_k^p)$ 
        \STATE $~~~~$$v_k^n = f_{\theta_t}(x_k^n)$ 
        \STATE $~~~~$\textcolor{gray}{\# embedding learning}
        \STATE $~~~~$$\mathcal{L} = \sum_{i=1}^{N} - log \frac{d(v_k^a, v_k^p) }{d(v_k^a, v_k^p) + d(v_k^a, v_k^n) + \sum_{j=1}^{K} d(v_i^a, B_j^a)}$
        \STATE update networks $f_{\theta_t}$ to minimize $\mathcal{L}$
        \STATE $~~~~$\textcolor{gray}{\# update memory bank}
        \STATE $~~~~$drop the earliest anchor feature in the memory bank $B$
        \STATE $~~~~$save anchor feature $v_k^a$ in the memory bank $B$
        \STATE $~~~~$\textcolor{gray}{\# weight-averaged}
        \STATE $f_{\theta_{t'}} = m f_{\theta_{t'}} + (1-m)f_{\theta_{t}} $
    \ENDFOR
    \STATE \textbf{return} encoder network $f_{\theta_{t'}}$
\end{algorithmic}
\end{algorithm}

\section{Experiments}
In this section, we first describe the datasets in \cref{sec:data}.
Then the implementation details and experimental settings are illustrated
in \cref{sec:id}.
The results of comparison with state-of-the-art methods are reported in \cref{sec:sota}.
Finally, a series of ablation studies are studied in \cref{sec:ablation_study}.

\subsection{Datasets}
\label{sec:data}
The self-supervised training are performed on three datasets: \textbf{UCF101} consists of 13,320 manually labeled videos from 101 action categories. It has three train/test splits and each split has around 9,500 videos for training and 3,700 videos for test.
\textbf{HMDB51} is a realistic and challenging dataset whose action categories mainly differ in motion rather than static poses. It consists of 6,766 manually labeled clips from 51 categories and three train/test splits.
\textbf{Kinetics} has 400 human action classes with more than 400 clips for each class. The validation set of Kinetics consists of about 20k video clips. 

\subsection{Implementation Details and Settings}
\label{sec:id}

\subsubsection{Networks}
We first experiment on traditional C3D \cite{tran2015learning}.
To further demonstrate the generality of our method, we also employ large networks include R3D \cite{hara2018can}, R(2+1)D \cite{tran2018closer} and I3D \cite{carreira2017quo} as baselines.
All these 3D backbone contain global avg pooling (GAP) layer before final fully-connected (FC) layer. 
During self-supervised pre-train, we replace GAP and FC layer with our Projection head for self-supervised learning which will be described in next subsection.

\subsubsection{Projection}
In this paper, we adopt four types of 3D CNN as the backbone encoder and each network has different dimension of the output. 
The Projection head of self-supervised learning on Kinetics is shown at Figure \ref{fig:projection}, which consist of spatio-temporal feature aggregation layers and non-linear MLP.
We adopt L2 normalization at the end of projection as same as \cite{wu2018unsupervised}.
Considering the massive parameters of both temporal convolutional layer and MLP, we replace the first part with global average pooling layer on small size HMDB51 and UCF101 to prevent overfitting.
Besides, we apply one fully-connected layer with fixed-dimensional output (128-D) rather than MLP after the first part.

\begin{figure}
  \centering
  \includegraphics[width=.8\linewidth]{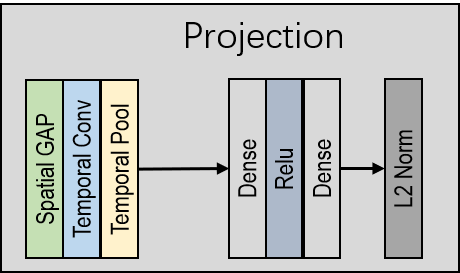}
  \caption{\textbf{Projection Head}: consists of spatiao-temporal feature fusion layers and MLP.}
  \label{fig:projection}
\end{figure}

\begin{table}[t]
    \centering
    {
    \caption{Comparison of our VTDL with state-of-the-art 2D/3D CNN based methods on UCF101 and HMDB51. The first line report the result of 2D CNN based method while the second line is based on 3D CNN.}
    \label{tab:sota_action_recognition_cmp}
    
    \begin{tabular}{ccc}
    \hline
    Method& UCF101(\%) & HMDB51(\%) \\
    \hline
    Shuffle\&Learn\cite{misra2016shuffle} \textcolor[rgb]{0,0,1}{[ECCV, 2016]} &50.2&18.1\\
    VGAN \cite{vondrick2016generating} \textcolor[rgb]{0,0,1}{[NeulPS, 2016]} &51.1 &-\\
    OPN \cite{lee2017unsupervised} \textcolor[rgb]{0,0,1}{[ICCV, 2017]} &56.3 &22.1 \\
    \hline
    ST-puzzle \cite{kim2019self} \textcolor[rgb]{0,0,1}{[AAAI, 2019]} &60.6 & 28.3\\
    Skip-Clip \textcolor[rgb]{0,0,1}{[ICCV, 2019]} &64.4 & -\\
    Statistical \cite{wang2019self2}
    \textcolor[rgb]{0,0,1}{[CVPR, 2019]} &- &32.6 \\
    Clip Order \cite{xu2019self}
    \textcolor[rgb]{0,0,1}{[CVPR, 2019]} &65.6 &28.4 \\
    VCP \cite{luo2020video} \textcolor[rgb]{0,0,1}{[AAAI, 2020]} & 68.5 & 32.5 \\
    \hline
    \textbf{VTDL} & 73.2 & 40.6 \\
    \textbf{VTDL} (Kinetics) & 75.5 & 43.2 \\
    \hline
    \end{tabular}
    }
\end{table}

\subsubsection{Evaluation}
The goal of our self-supervised task is to learn a temporal-discriminative representation with well generalization for downstream action recognition task.
For this purpose, the evaluation of our self-supervised method consists of two essential steps:

\textbf{Step 1: Self-supervised Pretrain.}
We use SGD as our optimizer. 
The SGD weight decay is 5e-4 and the SGD momentum is 0.9. 
For all datasets, we use a mini-batch in size of 64 on 4 GPUs, and an initial learning rate of 0.01.
The total epochs is 50 and the learning rate goes down to 1/10 every 10 epochs.
We train our method on split 1 of UCF101 and HMDB51, train set for kinetics.
Notice use all dataset as training set lead to better result.
The smoothing coefficient hyperparameter $m$ is 0.99 as default and $d$ is inner product as in \cite{he2019momentum}.
For each clip, we randomly sample 16 frames with temporal stride 4, and spatially resize these frames as 224 $\times$ 224 pixels. 
We also explore the influence of temporal stride in \cref{sec:ablation_study}.

\textbf{Step 2: Transfer of Learned Representations.}
After finishing the pretrain stage, we use our learned parameters to initialize the 3D CNNs for downstream action recognition, while the last fully connected layer is randomly initialized. 
During the fine-tuning and testing period, we follow the same protocol in \cite{carreira2017quo} to provide a fair comparison. 
We apply random spatial cropping and flip horizontal to perform data augmentation.

In \cref{sec:sota}, to make thorough comparison with state-of-the-art methods, we follow the common setting in \cite{kim2019self,wang2019self}, the input video clip is 16 consistence frames with size $112\times112$. 
The final accuracy is obtained by mean average of all these clips with slide window stride 8.
We pretrain on split 1 of UCF101 and fine-tune on all 3 splits of HMDB51/UCF101.

In \cref{sec:ablation_study}, to compare with ImageNet pretrained model and existing fully supervised methods, we follow common experiment setting \cite{carreira2017quo,wang2018non} and fine tune on 64 frame clip with spatial size 224 $\times$ 224. 
The final result is averaged by 10 clips that selected by sliding window with step 16.
We carry out experiments on UCF101 and HMDB51 and the backbone we used is the deep and powerful I3D \cite{carreira2017quo}.



\begin{table}
  \caption{Ablation study of temporal-discriminative loss on UCF101 and HMDB51.}
  \label{tab:ablation_cmp}
  \begin{tabular}{ccc}
    \hline
    Method&UCF101&HMDB51\\
    \hline
    Scratch & 63.3 & 26.2 \\
    Removing Inter-video Contrast & 68.2 &32.6 \\
    Removing Intra-video Contrast & 70.4 & 35.5\\
    Temporal Discriminative Loss &72.7&41.2 \\
    + Data Augmentation & \textbf{82.3} & \textbf{52.9}\\
  \hline
\end{tabular}
\end{table}


\begin{table}
    \caption{Ablation study of data augmentation for temporal triplet on UCF101 and HMDB51. BA means Basic Augmentation.}
    \label{tab:data_augmentataion_cmp}
    \begin{tabular}{cccccc}
    \hline
    BA &Video Cutout & Ext- Mix & Int- Mix & UCF101 & HMDB51 \\
    \hline
    -&-&-&-&72.7&41.2 \\
    \checkmark&-&-&-&74.3&43.7 \\
    \checkmark&\checkmark&-&-&75.4&44.6 \\
    \checkmark&-&\checkmark&-&77.4&46.3 \\
    \checkmark&-&-&\checkmark&81.2&51.5 \\
    \checkmark&\checkmark&\checkmark&-&78.5&47.6  \\
    \checkmark&-&\checkmark&\checkmark&81.9&52.4 \\
    \checkmark&\checkmark&\checkmark&\checkmark&\textbf{82.3}&\textbf{52.9} \\
    \hline
    \end{tabular}
\end{table}

\subsection{Comparison with State-of-the-art Methods}
\label{sec:sota}

In this experiment, we evaluate the proposed framework against the recent state-of-the-art 2D/3D CNN based self-supervised solution on the datasets HMDB51 and UCF101.
We report the average classification over 3 splits.
For fair comparison, all 3D CNN based method adopt same C3D as backbone.
The results are summarized in Table \ref{tab:sota_action_recognition_cmp}.
It can be noticed that 3D CNN based methods outperform 2D based a lot.

Besides, we can draw the following conclusions from the evaluation results:
\emph{i.} The performance of VTDL is 73.2\% on UCF101 and 40.6\% on HMDB51 with same experiment setting, which outperforms 3D CNN based solution by a large margin. The model pretrained with our VTDL beats the state-of-the-art from VCP \cite{luo2020video} by 4.7\% on UCF101 and 8.1 on HMDB51.
\emph{ii.} With preforming self-supervised pretrain with more large Kinetics, our VTDL leads to better results. The result prove that our self-supervised learning technique has a good generalization ability.

\subsection{Ablation Study}
\label{sec:ablation_study}

In this section, a series of ablation studies are first performed to understand the importance of each component of temporal-discriminative loss and data augmentation.
Next, we explore the relative performance improvement of backbone and two hyperparameters in VTDL: the time constant $\tau$ and the temporal sampling stride.
Finally, we study the Benefit of additional unlabeled data.

\begin{table}
  \caption{Comparison with different backbones on UCF101 and HMDB51.}
  \label{tab:newotk_cmp}
  \begin{tabular}{cccc}
    \hline
    Method& Pretrain &UCF101&HMDB51\\
    \hline
    C3D & -& 59.4 & 22.7 \\
    C3D & \textbf{VTDL}& 73.8 (14.6$\uparrow$) & 46.6 (27.9$\uparrow$)  \\
    R3D-50& -& 60.7 & 25.1 \\
    R3D-50 &\textbf{VTDL}& 78.4 (17.7$\uparrow$) & 49.1 (24$\uparrow$) \\
    I3D &-&63.3 & 26.2 \\
    I3D &\textbf{VTDL}&82.1 (18.8$\uparrow$) & \textbf{52.9} (26.7$\uparrow$) \\
    R(2+1)D-50& - &65.4 & 26.8\\
    R(2+1)D-50& \textbf{VTDL} &\textbf{84.9} (19.5$\uparrow$) & 52.5 (25.7$\uparrow$)\\
    \hline
\end{tabular}
\end{table}

\subsubsection{Temporal-Discriminative Loss} 
To demonstrate the effectiveness of each component in Temporal-Discriminative Loss, we train several baseline models without any data augmentation by removing inter-video contrast (item $\sum_{j \neq i}^{} d(v_i^a, v_j^a)$ in equation \ref{eq:4}) and intra-video contrast (item $d(v_i^a, v_i^n)$ in equation \ref{eq:4}).
Experiment results are reported in Table \ref{tab:ablation_cmp}. 
Based on these results, we make two observations:
\emph{i.}
Learning without inter-video contrast or intra-video contrast, the recognition performance compared with Scratch is consistently improved, which evidences that doing contrast on temporal triplet is a powerful way for video self-supervised learning.
\emph{ii.}
VTDL achieve 82.3\% on UCF101 and 52.9\% on HMDB51 when all the components are integrated, showing that using data augmentation to diversify temporal triplet is still a plus even with strong temporal-discriminative learning.

\begin{figure}
      \centering
    	\includegraphics[width=\linewidth]{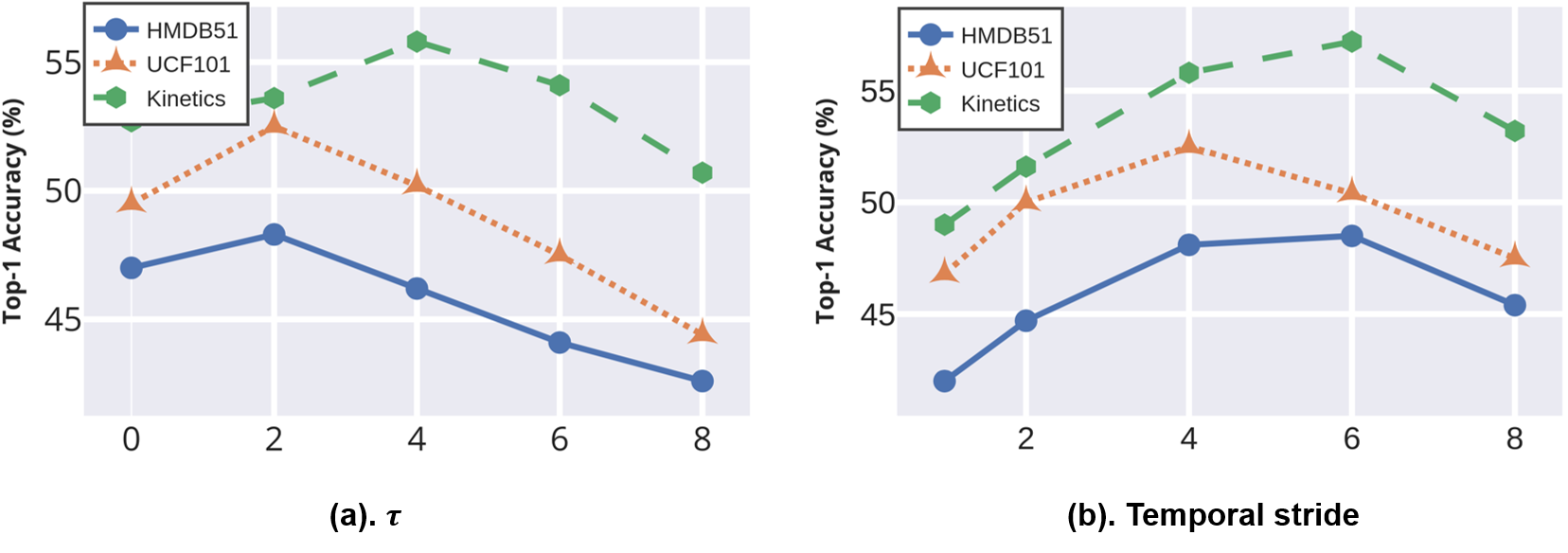}
    	\caption{Comparisons the performance of different hyperparameter on three datasets. 
    	\textbf{Left}: The timestamp $\tau$ between \textit{anchor} and \textit{negative}.
    	\textbf{Right}: The interval of temporal sampling.}
    	\label{fig:hyper}
\end{figure}

\begin{figure}
      \centering
    	\includegraphics[width=\linewidth]{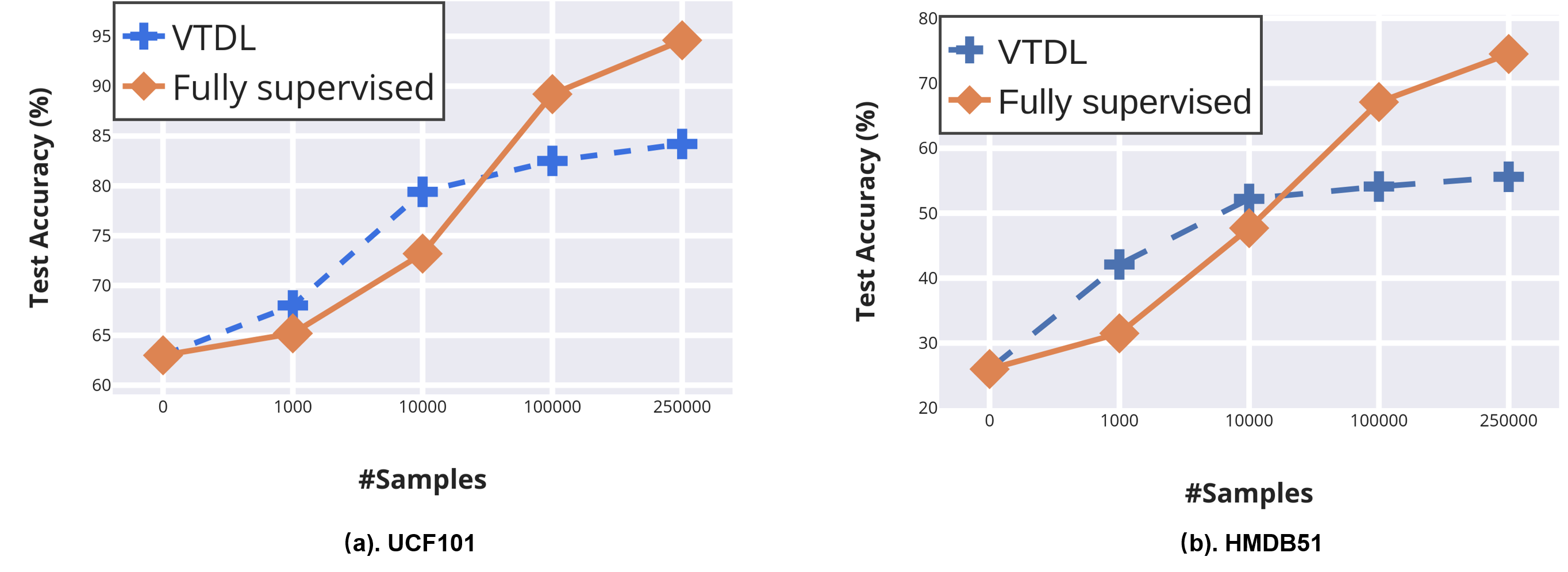}
    	\caption{The evaluation accuracy on UCF101 and HMDB51 with VTDL or fully supervised learning pretrain with Kinetics dataset.}
      \label{fig:sampleAcc}
\end{figure}

\begin{table}
  \caption{Comparison with fully-supervised learning on UCF101 and HMDB51.}
  \label{tab:i3d_cmp}
  \begin{tabular}{cccc}
    \hline
    Method& Pretrain Data&UCF101&HMDB51\\
    \hline
    Scratch & -& 63.3 & 26.2 \\
    ImageNet \cite{carreira2017quo} & $\sim$ {14m labeled image} & 84.1 & 47.6 \\
    UCF101 \cite{carreira2017quo}& $\sim$ {9k labeled video} & - & 49.7\\
    Kinetics \cite{carreira2017quo} & $\sim$ {220k labeled video}  & 94.6 & 74.5\\
    \hline
    \textbf{VTDL} (HMDB51) & $\sim$ {4k unlabeled video} & - & 47.3 \\
    \textbf{VTDL} (UCF101) & $\sim$ {9k unlabeled video} & 82.4 & 52.9 \\
    \textbf{VTDL} (Kinetics) & $\sim$ {220k unlabeled video} & \textbf{84.8} & \textbf{55.4} \\
  \hline
\end{tabular}
\end{table}

\subsubsection{Data Augmentation}
In this experiment we explore the effectiveness of four types data augmentation methods.
The results of this controlled experiment are shown in Table \ref{tab:data_augmentataion_cmp}. 
We observe Basic Augmentation improved the representation ability of Scratch on both datasets. 
Also, each component in Temporal Consistent Augmentation is effective and the Internal Mix achieve about 7\% improvement on UCF101 and 8\% on HMDB51.
The best result is obtained by the combination of all these augmentation techniques.
This result indicates that the idea of introducing extra spatial context is feasible and effective for video datasets to learn temporal-discriminative representation.

\subsubsection{Backbone.}
The relative model capacities of base feature extractor can impact the final transfer performance.
To study this behavior, we fix the training samples and vary the architecture of the feature extractor as follows: (a) shallow C3D. (b) R3D-50. (c) I3D. (d) R(2+1)D-50 model.
From Table \ref{tab:newotk_cmp}, we observe a consistent improvement of 2\% - 3\% when a higher capacity model was used as base CNNs.
This results is intuitive and indicates that a higher capacity base encoder yields richer visual features for VTDL and thus improves the transfer learning performance.

\begin{figure*}
  \centering
  \includegraphics[width=\linewidth]{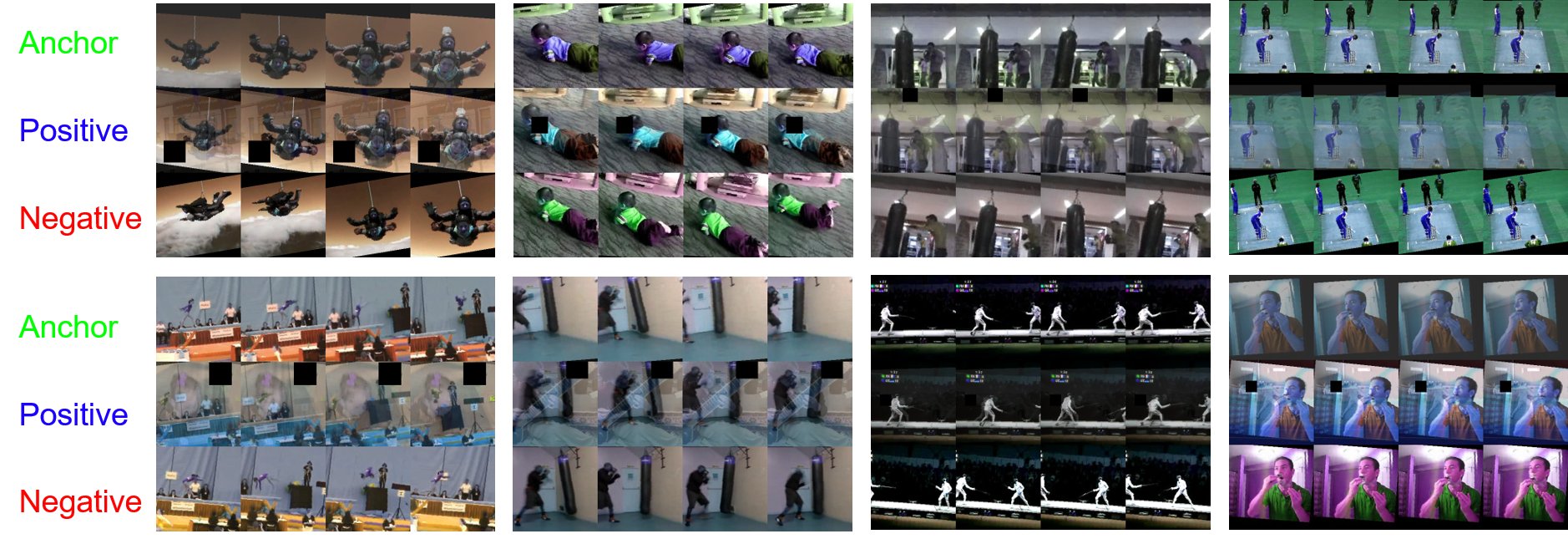}
  \caption{Visualization of the generated temporal triplets.}
  \label{fig:triplet_visualize}
\end{figure*}

\subsubsection{Hyperparameter.}
To study how varying the time constant $\tau$ and the temporal stride of input video clip effects VTDL, we fix other settings and vary the number of $\tau$ or temporal stride in Step 1.
The results of Step 2 on HMDB51 are reported in Figure \ref{fig:hyper}.
From Figure \ref{fig:hyper} (a), we observe that the best performance is obtained with $\tau=2$ on UCF101 and HMDB51 but $\tau=4$ on Kinetics, considering the average video length of Kinetics is several times larger than the others datasets.
Choosing appropriate time constant $\tau$ is important, the temporally closer negative clip share more common spatial context with a high probability as long as they are not very drastic action.
Figure \ref{fig:hyper} (b) shows the result of different temporal stride during training.
As the larger temporal stride, the top-1 accuracy performance (\%) is increasing, since bigger temporal stride for a fixed video implies reducing more appearance similarity between adjacent frames.
However for some short video, too large temporal stride may lead to unnatural actions, caused by that sampled timestamp suppress the length of whole video and loop from beginning again. 
For example, the performance drop a lot when temporal stride is large than 6.

\subsubsection{Benefit of additional unlabeled data}

In this section we explore the effect of using different amounts of unlabeled data and compare with fully supervised learning.
For fair comparison, unlabeled samples class-balanced selected from Kinetics are used to train a network under VTDL with fixed number of training epochs.
The relative performance on HMDB51 and UCF101 are shown in Figure \ref{fig:sampleAcc}. 
It suggests that using the more and diverse data is beneficial during self-supervised training, even each sample are seen fewer times.
We also observe that our VTDL leads to better performance than fully supervised learning when the datasets are less than 10 thousands. 
This experiment clearly demonstrate the utility of our approach in pretrain with limited training samples.

Furthermore, we also report the result of fully-supervised pretrained with different source of data in Table \ref{tab:i3d_cmp}. 
Our method not only suppress the ImageNet pretrained model but also UCF101 pretrained model.
The effectiveness on all datasets demonstrate our method is not sensitive to the domain of unlabeled dataset.
What's interesting is that even without new unlabeled data available, VTDL on the same dataset still outperforms Scratch by a large margin. 
It suggests that our VTDL can indeed learn some useful temporal information that are ignored by supervised learning with Cross-entropy loss.

\section{Temporal Triplet Visualization}


Figure \ref{fig:triplet_visualize} visualizes the generated temporal triplets by our method.
Specifically, we randomly select some videos with temporal stride 4 on UCF101.
We make two remarks for these generated triplets.
First, we notice that \textit{positive} has huge appearance difference with \textit{anchor} but maintain the same motion information, that is, we can also observe \textit{positive} describe the identical motion pattern with the \textit{anchor}.
Second, we observe that \textit{negative} contain similar background with \textit{anchor} but large difference in motion pattern.
In summary, temporal reasoning and insensitivity to appearance are crucial for video-based self-supervised learning based on the effectiveness of VTDL.

\section{Conclusion}
In this paper, we propose a general framework, Video-based Temporal-Discriminative Learning (VTDL), for self-supervised video representation learning, which is scalable and without any restrictions on model architectures, source of data, and forms of supervision.
The keys of boosting performance in VTDL are the online generated temporal triplet and Temporal Consistent Augmentation (TCA), which improve the data diversity of video dataset and help the model learn temporal-discriminative representation.
The experimental result shows that VTDL outperform prior self-supervised tasks and can match or improve the performance of networks trained on supervised data.
However, we intuitively regard different temporal window as different instance without considering complex characteristics of motion, such as periodic action, which could be improved by our further exploration.
Our code are also publicly available hoping boosting the development of the community.



%





\ifCLASSOPTIONcaptionsoff
  \newpage
\fi

\bibliographystyle{IEEEtran}
\bibliography{main.bib}

\begin{thebibliography}{10}
\providecommand{\url}[1]{#1}
\csname url@samestyle\endcsname
\providecommand{\newblock}{\relax}
\providecommand{\bibinfo}[2]{#2}
\providecommand{\BIBentrySTDinterwordspacing}{\spaceskip=0pt\relax}
\providecommand{\BIBentryALTinterwordstretchfactor}{4}
\providecommand{\BIBentryALTinterwordspacing}{\spaceskip=\fontdimen2\font plus
\BIBentryALTinterwordstretchfactor\fontdimen3\font minus
  \fontdimen4\font\relax}
\providecommand{\BIBforeignlanguage}[2]{{%
\expandafter\ifx\csname l@#1\endcsname\relax
\typeout{** WARNING: IEEEtran.bst: No hyphenation pattern has been}%
\typeout{** loaded for the language `#1'. Using the pattern for}%
\typeout{** the default language instead.}%
\else
\language=\csname l@#1\endcsname
\fi
#2}}
\providecommand{\BIBdecl}{\relax}
\BIBdecl

\bibitem{carreira2017quo}
J.~Carreira and A.~Zisserman, ``Quo vadis, action recognition? a new model and
  the kinetics dataset,'' in \emph{CVPR}, 2017, pp. 6299--6308.

\bibitem{wang2018non}
X.~Wang, R.~Girshick, A.~Gupta, and K.~He, ``Non-local neural networks,'' in
  \emph{CVPR}, 2018, pp. 7794--7803.

\bibitem{feichtenhofer2019slowfast}
C.~Feichtenhofer, H.~Fan, J.~Malik, and K.~He, ``Slowfast networks for video
  recognition,'' in \emph{ICCV}, 2019, pp. 6202--6211.

\bibitem{gidaris2018unsupervised}
S.~Gidaris, P.~Singh, and N.~Komodakis, ``Unsupervised representation learning
  by predicting image rotations,'' \emph{ICLR}, 2018.

\bibitem{DBLP:conf/iccv/DoerschGE15}
C.~Doersch, A.~Gupta, and A.~A. Efros, ``Unsupervised visual representation
  learning by context prediction,'' in \emph{ICCV}, 2015, pp. 1422--1430.

\bibitem{DBLP:conf/eccv/NorooziF16}
M.~Noroozi and P.~Favaro, ``Unsupervised learning of visual representations by
  solving jigsaw puzzles,'' in \emph{ECCV}, 2016, pp. 69--84.

\bibitem{DBLP:conf/cvpr/PathakKDDE16}
D.~Pathak, P.~Kr{\"{a}}henb{\"{u}}hl, J.~Donahue, T.~Darrell, and A.~A. Efros,
  ``Context encoders: Feature learning by inpainting,'' in \emph{CVPR}, 2016,
  pp. 2536--2544.

\bibitem{wu2018unsupervised}
Z.~Wu, Y.~Xiong, S.~X. Yu, and D.~Lin, ``Unsupervised feature learning via
  non-parametric instance discrimination,'' in \emph{CVPR}, 2018, pp.
  3733--3742.

\bibitem{ye2019unsupervised}
M.~Ye, X.~Zhang, P.~C. Yuen, and S.-F. Chang, ``Unsupervised embedding learning
  via invariant and spreading instance feature,'' in \emph{CVPR}, 2019, pp.
  6210--6219.

\bibitem{ZhuangZY19}
C.~Zhuang, A.~L. Zhai, and D.~Yamins, ``Local aggregation for unsupervised
  learning of visual embeddings,'' in \emph{ICCV}.\hskip 1em plus 0.5em minus
  0.4em\relax {IEEE}, 2019, pp. 6001--6011.

\bibitem{he2019momentum}
K.~He, H.~Fan, Y.~Wu, S.~Xie, and R.~Girshick, ``Momentum contrast for
  unsupervised visual representation learning,'' \emph{arXiv preprint
  arXiv:1911.05722}, 2019.

\bibitem{chen2020simple}
T.~Chen, S.~Kornblith, M.~Norouzi, and G.~Hinton, ``A simple framework for
  contrastive learning of visual representations,'' \emph{arXiv preprint
  arXiv:2002.05709}, 2020.

\bibitem{misra2016shuffle}
I.~Misra, C.~L. Zitnick, and M.~Hebert, ``Shuffle and learn: unsupervised
  learning using temporal order verification,'' in \emph{ECCV}.\hskip 1em plus
  0.5em minus 0.4em\relax Springer, 2016, pp. 527--544.

\bibitem{xu2019self}
D.~Xu, J.~Xiao, Z.~Zhao, J.~Shao, D.~Xie, and Y.~Zhuang, ``Self-supervised
  spatiotemporal learning via video clip order prediction,'' in \emph{CVPR},
  2019, pp. 10\,334--10\,343.

\bibitem{kim2019self}
D.~Kim, D.~Cho, and I.~S. Kweon, ``Self-supervised video representation
  learning with space-time cubic puzzles,'' in \emph{Proceedings of the AAAI
  Conference on Artificial Intelligence}, vol.~33, 2019, pp. 8545--8552.

\bibitem{doersch2015unsupervised}
C.~Doersch, A.~Gupta, and A.~A. Efros, ``Unsupervised visual representation
  learning by context prediction,'' in \emph{Proceedings of the IEEE
  International Conference on Computer Vision}, 2015, pp. 1422--1430.

\bibitem{hadsell2006dimensionality}
R.~Hadsell, S.~Chopra, and Y.~LeCun, ``Dimensionality reduction by learning an
  invariant mapping,'' in \emph{CVPR}, vol.~2.\hskip 1em plus 0.5em minus
  0.4em\relax IEEE, 2006, pp. 1735--1742.

\bibitem{fernando2017self}
B.~Fernando, H.~Bilen, E.~Gavves, and S.~Gould, ``Self-supervised video
  representation learning with odd-one-out networks,'' in \emph{CVPR}, 2017,
  pp. 3636--3645.

\bibitem{wang2019self}
J.~Wang, J.~Jiao, L.~Bao, S.~He, Y.~Liu, and W.~Liu, ``Self-supervised
  spatio-temporal representation learning for videos by predicting motion and
  appearance statistics,'' in \emph{CVPR}, 2019, pp. 4006--4015.

\bibitem{han2019video}
T.~Han, W.~Xie, and A.~Zisserman, ``Video representation learning by dense
  predictive coding,'' in \emph{ICCVW}, 2019, pp. 0--0.

\bibitem{luo2020video}
D.~Luo, C.~Liu, Y.~Zhou, D.~Yang, C.~Ma, Q.~Ye, and W.~Wang, ``Video cloze
  procedure for self-supervised spatio-temporal learning,'' \emph{AAAI}, 2020.

\bibitem{sermanet2018time}
P.~Sermanet, C.~Lynch, Y.~Chebotar, J.~Hsu, E.~Jang, S.~Schaal, S.~Levine, and
  G.~Brain, ``Time-contrastive networks: Self-supervised learning from video,''
  in \emph{ICRA}.\hskip 1em plus 0.5em minus 0.4em\relax IEEE, 2018, pp.
  1134--1141.

\bibitem{wang2018temporal}
L.~Wang, Y.~Xiong, Z.~Wang, Y.~Qiao, D.~Lin, X.~Tang, and L.~Van~Gool,
  ``Temporal segment networks for action recognition in videos,'' \emph{TPAMI},
  vol.~41, no.~11, pp. 2740--2755, 2018.

\bibitem{DBLP:conf/mm/RuizPBM17}
A.~H. Ruiz, L.~Porzi, S.~R. Bul{\`{o}}, and F.~Moreno{-}Noguer, ``3d cnns on
  distance matrices for human action recognition,'' in \emph{ACM MM}, 2017.

\bibitem{DBLP:journals/tmm/HouWSJ18}
J.~Hou, X.~Wu, Y.~Sun, and Y.~Jia, ``Content-attention representation by
  factorized action-scene network for action recognition,'' \emph{{IEEE} Trans.
  Multimedia}, vol.~20, no.~6, pp. 1537--1547, 2018.

\bibitem{DBLP:conf/mm/JiX00SH19}
Y.~Ji, F.~Xu, Y.~Yang, N.~Xie, H.~T. Shen, and T.~Harada, ``Attention transfer
  {(ANT)} network for view-invariant action recognition,'' in \emph{ACM MM},
  2019.

\bibitem{DBLP:conf/mm/LiuGQWL19}
Z.~Liu, G.~Gao, A.~K. Qin, T.~Wu, and C.~H. Liu, ``Action recognition with
  bootstrapping based long-range temporal context attention,'' in \emph{ACM
  MM}, 2019.

\bibitem{DBLP:conf/mm/0003ZWCLC19}
H.~Wu, Z.~Zha, X.~Wen, Z.~Chen, D.~Liu, and X.~Chen, ``Cross-fiber
  spatial-temporal co-enhanced networks for video action recognition,'' in
  \emph{ACM MM}, 2019.

\bibitem{hara2018can}
K.~Hara, H.~Kataoka, and Y.~Satoh, ``Can spatiotemporal 3d cnns retrace the
  history of 2d cnns and imagenet?'' in \emph{CVPR}, 2018, pp. 6546--6555.

\bibitem{tran2018closer}
D.~Tran, H.~Wang, L.~Torresani, J.~Ray, Y.~LeCun, and M.~Paluri, ``A closer
  look at spatiotemporal convolutions for action recognition,'' in \emph{CVPR},
  2018, pp. 6450--6459.

\bibitem{schroff2015facenet}
F.~Schroff, D.~Kalenichenko, and J.~Philbin, ``Facenet: A unified embedding for
  face recognition and clustering,'' in \emph{CVPR}, 2015, pp. 815--823.

\bibitem{1544882}
M.~{Blank}, L.~{Gorelick}, E.~{Shechtman}, M.~{Irani}, and R.~{Basri},
  ``Actions as space-time shapes,'' in \emph{ICCV}, vol.~2, 2005, pp.
  1395--1402 Vol. 2.

\bibitem{5196739}
A.~{Briassouli} and I.~{Kompatsiaris}, ``Robust temporal activity templates
  using higher order statistics,'' \emph{IEEE Transactions on Image
  Processing}, vol.~18, no.~12, pp. 2756--2768, 2009.

\bibitem{910878}
A.~F. {Bobick} and J.~W. {Davis}, ``The recognition of human movement using
  temporal templates,'' \emph{IEEE Transactions on Pattern Analysis and Machine
  Intelligence}, vol.~23, no.~3, pp. 257--267, 2001.

\bibitem{DBLP:journals/corr/cutout}
T.~Devries and G.~W. Taylor, ``Improved regularization of convolutional neural
  networks with cutout,'' \emph{arXiv preprint arXiv:1708.04552}, 2017.

\bibitem{weinberger2006distance}
K.~Q. Weinberger, J.~Blitzer, and L.~K. Saul, ``Distance metric learning for
  large margin nearest neighbor classification,'' in \emph{NeurlPS}, 2006, pp.
  1473--1480.

\bibitem{tarvainen2017mean}
A.~Tarvainen and H.~Valpola, ``Mean teachers are better role models:
  Weight-averaged consistency targets improve semi-supervised deep learning
  results,'' in \emph{NeurlPS}, 2017, pp. 1195--1204.

\bibitem{tran2015learning}
D.~Tran, L.~Bourdev, R.~Fergus, L.~Torresani, and M.~Paluri, ``Learning
  spatiotemporal features with 3d convolutional networks,'' in \emph{ICCV},
  2015, pp. 4489--4497.

\bibitem{vondrick2016generating}
C.~Vondrick, H.~Pirsiavash, and A.~Torralba, ``Generating videos with scene
  dynamics,'' in \emph{NeurlPS}, 2016, pp. 613--621.

\bibitem{lee2017unsupervised}
H.-Y. Lee, J.-B. Huang, M.~Singh, and M.-H. Yang, ``Unsupervised representation
  learning by sorting sequences,'' in \emph{ICCV}, 2017, pp. 667--676.

\bibitem{wang2019self2}
J.~Wang, J.~Jiao, L.~Bao, S.~He, Y.~Liu, and W.~Liu, ``Self-supervised
  spatio-temporal representation learning for videos by predicting motion and
  appearance statistics,'' in \emph{CVPR}, 2019, pp. 4006--4015.

\end{thebibliography}




\begin{IEEEbiographynophoto}{Jinpeng Wang}
received the B.E. degree in Software Engineering from Sun Yat-sen University, Guangzhou, China, in 2017. He is currently a graduate student in School Of Electronics And Information Technology, Sun Yat-sen University, Guangzhou, China. His current research interests include biometrics, video understanding.
\end{IEEEbiographynophoto}

\begin{IEEEbiographynophoto}{Yiqi Lin}
received the bachelor's degree in software engineering from Sun Yat-Sen University in 2019. He will be a M.S. student in the School of Data and Computer Science in Sun Yat-Sen University. His research interests include computer vision and machine learning.
\end{IEEEbiographynophoto}

\begin{IEEEbiographynophoto}{Andy J. Ma}
received his B.Sc. and M.Sc. degrees in applied mathematics from Sun Yat-sen University, respectively, and the Ph.D. degree from the Department of Computer
Science, Hong Kong Baptist University. He worked as a Post-Doctoral Fellow at Rutgers University and Johns Hopkins University. Now, he is an associate professor at Sun
Yat-sen University. His current research interests focus on developing machine learning algorithms for intelligent video surveillance and medical applications.
\end{IEEEbiographynophoto}

\begin{IEEEbiographynophoto}{Pong C. Yuen}
received the B.Sc.
(Hons.) degree in electronic engineering from City Polytechnic of Hong Kong, in 1989, and the Ph.D. degree in electrical and electronic engineering from The University of Hong Kong, in 1993. He joined the Hong Kong Baptist University, in 1993, where he is currently a Professor and the Head of the Department of Computer Science. Dr. Yuen spent a six-month sabbatical leave with The University of Maryland Institute for Advanced Computer Studies, University of Maryland at College Park, in 1998. From 2005 to 2006, he was a Visiting Professor with the GRAphics, VIsion and Robotics Laboratory, INRIA, Rhone Alpes, France. He was the Director of the Croucher Advanced Study Institute on Biometric Authentication in 2004, and the Director of Croucher ASI on Biometric Security and Privacy in 2007. He was a recipient of the University Fellowship to visit The University of Sydney in 1996. Dr. Yuen has been actively involved in many international conferences as an Organizing Committee and/or a Technical Program Committee Member. He was the Track Co-Chair of the International Conference on Pattern Recognition in 2006 and the Program Co-Chair of the IEEE Fifth International Conference on Biometrics: Theory, Applications and Systems in 2012. He is an Editorial Board Member of Pattern Recognition and the Associate Editor of the IEEE TRANSACTIONS ON INFORMATION FORENSICS AND SECURITY, and the SPIE Journal of Electronic Imaging. He is also serving as a Hong Kong Research Grants Council Engineering Panel Member. Dr. Yuen’s current research interests include video surveillance, human face recognition, biometric security, and privacy.
\end{IEEEbiographynophoto}




\end{document}